\documentclass{article}
\usepackage{ijcai26}
\usepackage{times}
\usepackage{soul}
\usepackage{url}
\usepackage[hidelinks]{hyperref}
\usepackage[utf8]{inputenc}
\usepackage[small]{caption}
\usepackage{graphicx}
\usepackage{amsmath}
\usepackage{amsthm}
\usepackage{booktabs}
\usepackage{algorithm}
\usepackage{algorithmic}
\usepackage{multirow}
\usepackage[switch]{lineno}
\usepackage{float}
\usepackage{placeins}
\usepackage{bm}
\usepackage{amsfonts}
\usepackage{natbib}
\usepackage{placeins}

\title{SymFold: Synergizing Evolutionary and Structural
Priors for Accurate Protein Inverse Folding}

\author{
Handong Wang$^{1,2}$,
Jiaxin Qi$^{1}$,
Baisheng Lai$^{1,2}$,
Jianqiang Huang$^{1,2}$
\affiliations
$^{1}$Computer Network Information Center, Chinese Academy of Sciences\\
$^{2}$University of Chinese Academy of Sciences
\emails
wanghandong24@mails.ucas.ac.cn,
jxqi@cnic.cn,
bslai@cnic.cn,
jqhuang@cnic.cn
}

\begin{document}

\maketitle
\begin{abstract}

Protein inverse folding aims to recover amino acid sequences for a given 3D protein structure, underpinning broad applications such as enzyme engineering and drug discovery.Current methods often follow a serial pipeline, in which a structure encoder predicts a coarse sequence, which is then refined by protein language models (PLMs). However, because PLMs only perform post-hoc sequence edits, the refinement is bounded by the quality of upstream predictions.Thanks to recent multimodal protein language models (MPLMs), we could directly encode structure to generate sequences with pretrained structural knowledge, but we observe that they are not effective for inverse folding. Therefore, we introduce a symmetric dual-path architecture that both leverages PLMs for pretrained sequence evolution knowledge and MPLMs for pretrained structural knowledge to iteratively guide protein sequence generation.Through extensive experiments across standard protein inverse folding benchmarks, our method achieves state-of-the-art performance, surpassing prior approaches, and ablation studies validate the rationale of our symmetric design, revealing a promising direction for the community.
\end{abstract}

\section{Introduction}

Protein sequence and structure are two sides of the same coin. The protein sequence specifies the order of amino acids (residues), and once known, the corresponding protein can be chemically synthesized \cite{protein_design_summary1,protein_design,protein_design_summary3}. Conversely, protein sequences naturally fold into higher-order three-dimensional (3D) structures, and distinct structures typically confer distinct functions \cite{protein_design_summary2}. Mapping between sequence and structure in both directions is therefore a central theme in protein science \cite{seq_structure_relation}. The forward direction is the well-known protein folding task, which predicts structure (and hence function) from sequence \cite{alphafold2,rosettafold,alphafold3}. In this work, we focus on the opposite direction, \textit{protein inverse folding}, which aims to design an amino acid sequence that will stably fold into a specified 3D structure \cite{proteinmpnn,pifold,gradeif,carbordesign}, enabling applications such as enzyme engineering and drug discovery. For example, to add a new binding site on the surface of an existing enzyme, researchers first design the local 3D structure required for binding, then use inverse folding methods to derive candidate sequences for synthesis and experimental validation.

\begin{figure*}[t]
    \centering
    \includegraphics[width=\linewidth]{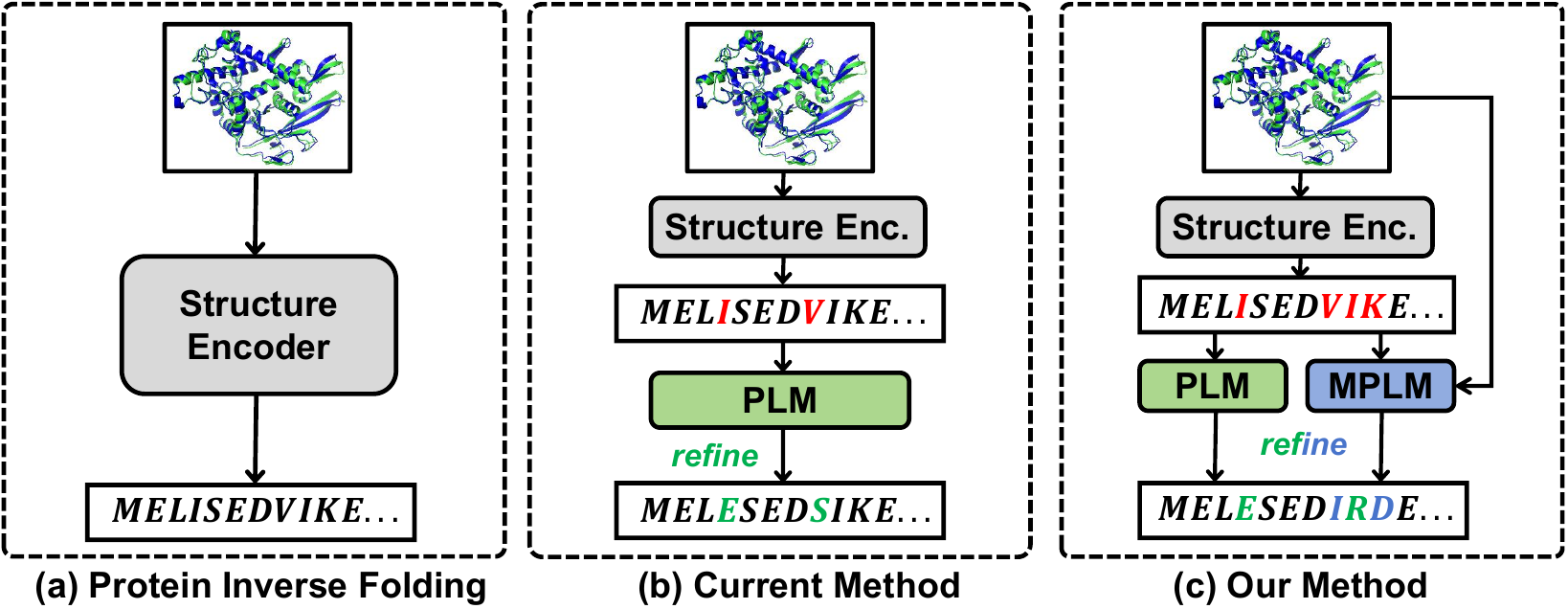}
    \caption{Comparison of various inverse folding frameworks: (a) The baseline model is trained from scratch without utilizing any pretrained knowledge to directly predict amino acid sequence from the protein structure. (b) PLM-based methods refine the output sequence of the structure encoder with pretrained sequence knowledge. (c) Our proposed SymFold symmetrically utilizes the pretrained knowledge of sequence and structure to refine the coarse sequence.}
    \label{fig:figure1}
\end{figure*}

Traditional protein inverse folding methods directly encode the 3D structure and perform position-wise amino acid (residue) classification (one-to-one mapping), as shown in Figure~\ref{fig:figure1}(a). Constrained by dataset scale and model capacity, their performance is limited. Besides, since the structure encoder ignores sequence context, the resulting sequences are usually biologically unreasonable. With the advent of large pretrained protein sequence models, i.e., protein language models (PLMs)~\cite{esm1b,esm2,progen,prottrans,proteinbert}, recent methods~\cite{kwdesign,bridge-if} often append a post-hoc refinement stage that uses PLMs to revise the output of the structure encoder. For example, as shown in Figure~\ref{fig:figure1}(b), if the encoder proposes ``\dots LISE\dots'' but assigns low confidence to ``I'', the PLM could leverage pretrained sequence knowledge to decide whether ``I'' should be replaced by another residue, conditioning on the context of ``I'', thereby improving sequence plausibility. However, such refinement is decoupled from the original structural evidence and is inherently upper-bounded by the first-stage output: when the proposed sequence already violates structural constraints, PLM edits may further degrade structural compatibility, compounding errors for inverse folding.

To bridge the gap between pretrained knowledge and given structures during refinement, we find that recent large pretrained multimodal protein models, i.e., multimodal protein language models (MPLMs)~\cite{esm3,dplm2,mplm_benchmark}, can directly map 3D structures to sequences and are naturally suited to the inverse folding task, even in a zero-shot manner. However, in practice, we find that MPLMs used in isolation are ineffective for this task, perhaps due to structural distribution shifts between pretraining and this task or their limited sequence modeling capacity. Therefore, we introduce \textbf{SymFold}, a symmetric dual-path framework that harmonizes PLMs and MPLMs in parallel to provide complementary evolutionary and structural priors for sequence generation.Specifically, as shown in Figure~\ref{fig:figure1}(c), our method first follows the traditional approach to derive a coarse sequence via a structure encoder, then we employ PLMs to provide pretrained sequence context knowledge, and meanwhile, leverage MPLMs to provide pretrained structural knowledge to refine the proposed sequence. Unlike the traditional uses of PLMs, our symmetric dual-path refinement simultaneously accounts for contextual dependencies and the residue-specific positional information, thereby addressing the refinement gap.

Furthermore, because the dependence on structure and sequence context varies across residues, we design an  Adaptive Synergistic Fusion (ASF) module that dynamically integrates structure cues, context dependencies, and structural priors so that the refinement can maintain compatibility with residue-specific biochemical characteristics. We also introduce a self-correction iterative training strategy to align training with the inference routine. A structure encoder first proposes a coarse sequence, PLMs and MPLMs perform an initial refinement, and then stochastically mix the outputs with the coarse sequence before the final refinement stage. This iterative procedure mirrors the test-time iterative prediction manner and reduces the mismatch between training and testing.

On standard inverse folding benchmarks, including the widely used CATH~\cite{cath}, TS50, and TS500~\cite{ts}, our method sets new state-of-the-art performance. Extensive ablations corroborate our findings and validate the proposed design, revealing the limitations of stand-alone MPLMs, the benefit of our symmetric dual-path architecture, and the effectiveness of our adaptive integration and self-correction modules. Our main contributions are summarized as follows:

\begin{itemize}
\item We diagnose the shortcomings of post-hoc refinement in current methods and argue that effective refinement must incorporate raw 3D evidence.
\item We introduce MPLMs into the inverse folding and empirically show that MPLMs alone are insufficient for this task.
\item We propose SymFold, which combines pretrained sequence and structural knowledge via Adaptive Synergistic Fusion and self-correction for sequence generation, achieving SOTA performance across multiple benchmarks and informing future directions.
\end{itemize}

\section{Related Work}
\textbf{Protein Inverse Folding}.
Early approaches in protein inverse folding primarily relied on physics-based methods such as Rosetta Design~\cite{rosetta}, which searched for low-energy sequences compatible with target backbones via sampling strategies like Monte Carlo simulation~\cite{advances}. While foundational, these methods were limited by approximate energy functions and high computational costs. The advent of deep learning~\cite{vae,gan,attention,gradient,bert,diffusion} enabled the use of geometric deep learning, where protein structures are modeled as graphs and processed with GNNs or equivariant networks to map structures to sequences. Representative works include GraphTrans~\cite{struct-gnn}, GVP~\cite{gvp}, and ProteinMPNN~\cite{proteinmpnn}, which achieved strong sequence recovery through optimized graph representations and message passing. PiFold~\cite{pifold} further improved efficiency with non-autoregressive predictions. Despite these advances, purely geometric models rely on limited structure-sequence pairs and struggle to capture evolutionary covariation present in massive sequence databases. To address this, such as LM-Design~\cite{lm-design}, Knowledge-Design~\cite{kwdesign}, and Bridge-IF~\cite{bridge-if}, couple protein language models (PLMs)~\cite{esm1b,esm2} with geometric models. Typically, these methods follow a serial refinement pipeline, where a PLM refines the initial output of a structure encoder. While effective, this serial pipeline suffers from decoupled refinement where PLMs lack access to structural constraints. In contrast, our SymFold introduces a symmetric dual-path architecture that synergizes evolutionary knowledge from PLMs with structural priors from MPLMs, ensuring that the sequence refinement remains grounded in both biological evolution and geometric constraints.

\textbf{Multimodal Protein Language Models (MPLMs)}.  
Recent MPLMs aim to unify sequences and structures under a single framework. ESM-3~\cite{esm3} embodies this direction by jointly modeling sequences, atomic coordinates, and functional annotations with Transformers, though its discretization of continuous coordinates constrains its direct application to inverse folding. Other notable MPLMs include DPLM-2~\cite{dplm2}, a diffusion-based framework capable of handling discrete and structural modalities, and Evolla~\cite{Evolla}, an 80B-parameter model that integrates protein-specific knowledge for design and functional tasks. These models highlight the promise of MPLMs for generative protein tasks, though challenges remain in fully exploiting them for inverse folding. In this study, we harness the structure-aware capabilities of MPLMs through our proposed framework, further unlocking their potential for such tasks.

\section{Method}

\subsection{Preliminaries}
\noindent\textbf{Baseline}. Protein inverse folding seeks to recover an amino acid sequence \(\bm{s} \!=\! (s_1, s_2, \ldots, s_\ell)\)  from a given protein structure \(\bm{x} \!=\! (x_1, x_2, \ldots, x_\ell)\), where \(s_i\!\in\!\mathcal{S}\) denotes the residue identity at position \(i\), $\mathcal{S}$ is the set of amino acids, and \({x}_i\) denotes the atomic coordinates for residue \(i\). The one-hot encoding of $s_i$ is $\bm{y}_{i}\!\in\!\{0,1\}^{|\mathcal{S}|}$ with components $(\bm{y}_{i})_a \!=\! \mathbb{I}\{a \!=\! s_{i}\}, a\in\mathcal{S}$, where $\mathbb{I}$ is the indicator function. 
Given training set \(\mathcal{D}\!=\!\{(\bm{x},\bm{y})\}\), the baseline methods learn a structure encoder $f_{e}$ producing per-position prediction (with entire structure of the protein), and the per-sample training objective could be written as sequence-averaged cross-entropy loss:
\begin{align}
\label{eq:baseline}
\mathcal{L}_\text{base} &=-\frac{1}{\ell}\sum_{i=1}^\ell  \bm{y}_i \cdot \log \frac{\exp\!\left(f_e(\bm{x})_i\right)}{\sum\nolimits_{k} \exp\!\left(f_e(\bm{x})_{i,k}\right)},
\end{align}
where $\exp(\cdot)$ is the exponential, $f_e(\bm{x})_i\!\in\!\mathbb{R}^{|\mathcal{S}|}$ is the predicted logits for $s_i$, and $k$ indexes classes of $\mathcal{S}$. 
In training, one minimizes the empirical mean of \(\mathcal{L}_\text{base}\) over all samples in the dataset \(\mathcal{D}\).

\noindent\textbf{PLM-based Method}. To incorporate sequence evolutionary knowledge, recent methods use a protein language model (PLM) $f_{s}$ to refine the predictions of an optimized structure encoder $f_e^*$. Specifically, they first train a structure encoder $f_e$ via Eq.~(\ref{eq:baseline}), and then finetune $f_{s}$ conditioning on the outputs of $f_e^*$, to conduct post hoc refinement. The sequence-level training objective is:
\begin{align}
\label{eq:plm}
\mathcal{L}_\text{plm} &=-\frac{1}{\ell}\sum_{i=1}^\ell  \bm{y}_i \cdot \log \frac{\exp\!\left(f_s(\hat{\bm{s}})_i\right)}{\sum\nolimits_{k} \exp\!\left(f_s(\hat{\bm{s}})_{i,k}\right)},\; \hat{\bm{s}}_j\!=\!\operatorname*{argmax}_k{f_e^*(\bm{x})_{j,k}},
\end{align}

where $\hat{\bm{s}}$ is the output of the optimal (and frozen) structure encoder $f_e^{*}$, the $\operatorname*{argmax}$ is taken over the class index $k$ for each residue $j$, and $f_s$ is trainable during finetuning. Note that, since no structure information $\bm{x}$ is available to $f_s$ during refinement, the post-hoc editing is inherently limited by $f_e^{*}$.

\begin{figure*}[t]
\centering
\includegraphics[width=1.0\linewidth]{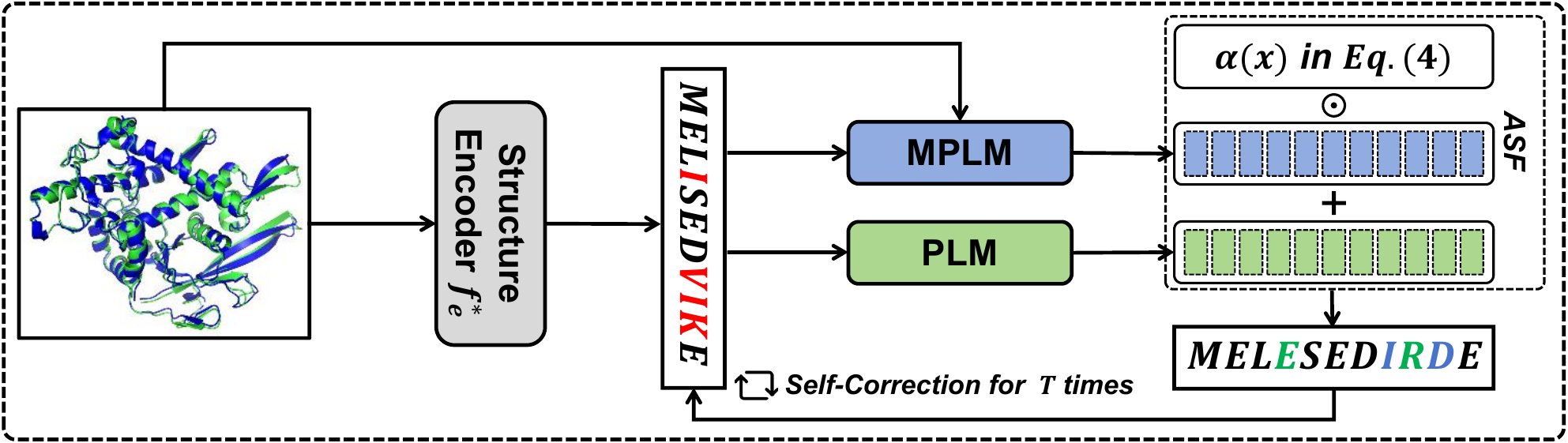}
\caption{\textbf{Overview of SymFold}. SymFold employs a symmetric refinement framework integrating structural priors from MPLMs and evolutionary priors from PLMs. Starting with a pre-trained structure encoder $f_e^*$, initial predictions are refined through parallel structure and sequence branches. The Adaptive Synergistic Fusion module combines these signals using structure-aware coefficients $\alpha(\bm{x})$ in Eq.(~\ref{eq:fusion}), $\odot$ denotes element-wise product, followed by iterative self-correction to align training with inference.
}
\label{fig:our_method}
\end{figure*}

\subsection{Our Method --- SymFold}

\noindent\textbf{Overview}.
To address the limitations of structure-ignorant refinement in PLM-based methods, we introduce SymFold, a symmetric dual-path architecture that couples structural priors from multimodal protein language models (MPLMs) with evolutionary priors from PLMs for stronger refinement. As shown in Figure~\ref{fig:our_method}, we first follow standard practice to train a structure encoder $f_e$ by Eq.~(\ref{eq:baseline}). We then encode structure knowledge via an MPLM and sequence knowledge via a PLM, and use them to adjust the sequence predictions through an \emph{Adaptive Synergistic Fusion} module. A \emph{self-correction} scheme is further proposed to align training and inference to mitigate exposure bias and reduce error accumulation. Together, these components enable a more harmonious and effective use of pretrained knowledge for protein inverse folding.

\noindent\textbf{Symmetric Refinement}. Following PLM-based practice, we first train a structure encoder via Eq.~(\ref{eq:baseline}) to obtain $f_e^{*}$. We then incorporate \emph{sequence} priors using a PLM $f_s$ based on the predicted sequence $\hat{\bm{s}}=\operatorname*{argmax}{f_e^*(\bm{x})}$ and incorporate \emph{structure} priors using an MPLM $f_m$ conditioned on $\bm{x}$. These two signals are fused by a symmetric module $f_{\text{sym}}$, and the per-sample training objective of our symmetric refinement is:
\label{method3}
\begin{align}
\label{eq:loss_ours}
\mathcal{L}_\text{ours} &=-\frac{1}{\ell}\sum_{i=1}^\ell  \bm{y}_i \cdot \log \frac{\exp\!\left(f_{sym}(f_s,\hat{\bm{s}},f_m,\bm{x})_i\right)}{\sum\nolimits_{k} \exp\!\left(f_{sym}(f_s,\hat{\bm{s}},f_m,\bm{x})_{i,k}\right)},
\end{align}
where $\exp(\cdot)$ is the exponential function, $\cdot$ is the dot product, $k$ indexes classes of $\mathcal{S}$. To effectively fuse pretrained knowledge from both paths, we introduce an Adaptive Synergistic Fusion Module that dynamically weights and blends their complementary cues. In addition, because inference aggregates more information sources, we propose a Self-Correction strategy that incrementally incorporates the dual paths step by step, thereby mitigating the train–test discrepancy.

\noindent\textbf{Adaptive Synergistic Fusion}. Unlike PLM-based refinement, which derives priors solely from sequence context, our symmetric refinement jointly leverages sequence context and structure signals. Since some residues are tightly constrained by local geometry while others depend more on long-range sequence dependencies, we design a residue-wise Adaptive Synergistic Fusion module that dynamically integrates these complementary cues:

\label{sec:fusion}
\begin{align}
\label{eq:fusion}
f_{sym}(f_s,\hat{\bm{s}},f_m,\bm{x})_i=f_s(\hat{\bm{s}})_i + \alpha(\bm{x})_i \cdot f_m(\bm{x},\hat{\bm{s}})_i,
\end{align}
where $\hat{\bm{s}}$ denotes the sequence predicted by the frozen optimal structure encoder $f_e^{*}$, $\cdot$ is the dot product, and $f_s$ and $f_m$ could be fine-tuned during training. The residue-wise structure-aware coefficient $\alpha(\bm{x})$ controls how strongly residue $i$ is influenced by the corresponding structure and could be realized as
$\alpha(\bm{x}) = \bm{w}^\top f_e^{*}(\bm{x})$,
where $\bm{w}$ is a trainable projection vector.

 
\noindent\textbf{Self-Correction}. Current PLM-based methods typically perform iterative generation at inference, feeding the predicted sequence back into the PLM for several rounds, which induces a mismatch between one-step training and multi-step inference. Under our symmetric framework, this discrepancy can be amplified since multiple knowledge sources are fused at each step. To address this, we introduce a Self-Correction strategy that unrolls the symmetric refinement during training to mirror the iterative inference procedure. The iterative process could be written as:

\begin{align}
\label{eq:sc}
\hat{\bm{s}}^{(0)} &= \operatorname*{argmax} f_e^{*}(\bm{x})\\
\hat{\bm{s}}^{(t)}_i &= \operatorname*{argmax}_k f_{sym}(f_s,\hat{\bm{s}}^{(t-1)},f_m,\bm{x})_{i,k},\,
\end{align}
where $t=1,\ldots,T$ is iterative step and $T$ is the total step. During training, we compute the loss using the refined prediction at $T$ for the loss calculation.

\section{Experiments}
\subsection{Experimental Protocol}
\textbf{Datasets.} We train our model on  CATH4.2 and CATH4.3. The CATH4.2 dataset consists of 18,024 proteins for training, 608 proteins for validation, and 1,120 proteins for testing, following the same data splitting as GraphTrans~\cite{struct-gnn}. The CATH4.3 dataset includes 16,153 structures for training, 1,457 for validation, and 1,797 for testing, following the same data splitting as ESMIF~\cite{ESMIF}.

For comprehensive evaluation, we assess our model on multiple protein structure datasets. We test on the CATH4.2 test set with 1,120 proteins and the CATH4.3 test set with 1,797 proteins to measure performance on protein folds similar to those seen during training. Additionally, we evaluate on the TS50 and TS500 datasets, comprising 50 and 500 proteins respectively as established by~\cite{ts}. To examine our model's generalization capabilities, we further test on the challenging CASP15~\cite{casp15} and CASP16~\cite{casp16} monomeric tertiary structure targets, which include 45 and 50 proteins respectively and represent novel protein structures. The specific protein identifiers and their official release times for CASP15 and CASP16 are provided in Appendix to facilitate reproducibility and cross-study comparison.

\textbf{Baseline Models and Evaluation Metrics.} We compare SymFold with recent graph-based models (StructGNN, GraphTrans~\cite{struct-gnn}, GCA~\cite{gca}, GVP~\cite{gvp}, GVP-large~\cite{ESMIF}, AlphaDesign~\cite{alphadesign}, ESM-IF~\cite{ESMIF}, ProteinMPNN~\cite{proteinmpnn},PiFold~\cite{pifold},GraDe-IF~\cite{gradeif}), and PLM-based optimized models (LM-Design~\cite{lm-design}, Knowledge-Design~\cite{kwdesign}, Bridge-IF~\cite{bridge-if}). We report perplexity and median recovery rate to assess performance(computation details in Appendix), with evaluations on the CATH dataset divided into three protein types: Short proteins (length $\leq$ 100), Single-chain proteins (with only 1 chain in CATH), and All proteins.

\textbf{Implementation Details.} SymFold employs a frozen pre-trained PiFold encoder for  fundamental structural representation, while only the Adaptive Synergistic Fusion module is fully fine-tuned. As language modeling backbones, we use the open-source ESM-3 (1.4B, MPLM) and its co-trained counterpart ESM-C (600M, PLM) \cite{esm3}. LoRA \cite{lora} is applied solely to MPLM and PLM with rank $r=8$, scaling factor $\alpha=32$, and dropout $0.1$, leading to $\sim$0.1\% trainable parameters in total (further configuration details are provided in the Appendix). Training is performed on a single NVIDIA A800 GPU with batch size 4, cosine learning rate scheduling, and typically converges within 5 epochs. For the self-correction training strategy, we set $T=2$ (more details about the inference phase setting of $T$ are in Appendix). All results are reported based on this final configuration.

\subsection{Results and Analysis}
Through the following Q\&A, we provide in-depth discussions of the experimental results for protein inverse folding, offering a comprehensive analysis of our SymFold model's performance across various benchmarks. We address key questions regarding performance comparisons, architectural innovations, generalization capabilities, ablation studies, and biological plausibility to systematically evaluate our approach's contributions to the field.

\textbf{Q1. How does SymFold perform on multiple  benchmarks?}

As shown in Table~\ref{tab:cath_results}, SymFold achieves state-of-the-art performance across all evaluated benchmarks. On CATH4.2, it reaches a 63.11\% overall sequence recovery rate, outperforming the best previous PLM-based method, Knowledge-Design, by 2.34\% and exceeding structure-only methods such as PiFold by 11.45 percentage points. The gains are especially notable for diverse protein categories: recovery on short proteins (length $\leq 100$) reaches 50.00\%, an improvement of 5.34\% over Knowledge-Design, while performance on single-chain proteins rises to 55.45\%, 6.49\% higher than Bridge-IF. These results highlight SymFold’s effectiveness in leveraging both sequence and structural information, particularly for structurally constrained proteins. On CATH4.3, the model maintains its advantage with a recovery rate of 62.23\%, showing consistent improvements across datasets. Furthermore, SymFold achieves the lowest perplexity across all protein categories, indicating more confident and accurate sequence predictions.

\begin{table*}[t]
\centering
\small
\setlength{\tabcolsep}{16pt}
\caption{Results comparison on the CATH4.2 and CATH4.3 datasets. Some benchmarked results are quoted from~\citep{kwdesign,bridge-if}. 
Perplexity ($\downarrow$) indicates sequence prediction uncertainty, where lower values are better; Recovery (\%$\uparrow$) measures sequence accuracy, where higher is better. The best and suboptimal results are labeled with bold and underline.}
\label{tab:cath_results}
\begin{tabular}{clcccccc}
\toprule
& \multirow{2}{*}{Model} & \multicolumn{3}{c}{Perplexity $\downarrow$} & \multicolumn{3}{c}{Recovery \% $\uparrow$} \\
\cmidrule(lr){3-5} \cmidrule(lr){6-8}
& & Short & Single-chain & All & Short & Single-chain & All \\
\midrule

\multirow{12}{*}{\rotatebox{90}{\textbf{CATH4.2}}} 
& StructGNN & 8.29 & 8.74 & 6.40 & 29.44 & 28.26 & 35.91 \\
& GraphTrans & 8.39 & 8.83 & 6.63 & 28.14 & 28.46 & 35.82 \\
& GCA & 7.09 & 7.49 & 6.05 & 32.62 & 31.10 & 37.64 \\
& GVP & 7.23 & 7.84 & 5.36 & 30.60 & 28.95 & 39.47 \\
& AlphaDesign & 7.32 & 7.63 & 6.30 & 34.16 & 32.66 & 41.31 \\
& ProteinMPNN & 6.21 & 6.68 & 4.61 & 36.35 & 34.43 & 45.96 \\
& PiFold & 6.04 & 6.31 & 4.55 & 39.84 & 38.53 & 51.66 \\
& GraDe-IF & 5.49 & 6.21 & 4.35 & 45.27 & 42.77 & 52.21 \\
& LM-Design & 5.66 & 5.52 & 4.01 & \underline{46.84} & 48.63 & 56.63 \\
& Knowledge-Design & \underline{5.48} & 5.16 & \underline{3.46} & 44.66 & 45.45 & \underline{60.77} \\
& Bridge-IF & 5.68 & \underline{5.06} & 3.83 & 43.86 & \underline{48.96} & 58.59 \\
\cmidrule(lr){2-8}
& \textbf{SymFold} & \textbf{4.69} & \textbf{4.01} & \textbf{3.23} & \textbf{50.00} & \textbf{55.45} & \textbf{63.11} \\

\midrule

\multirow{9}{*}{\rotatebox{90}{\textbf{CATH4.3}}} 
& GVP-large & 7.68 & 6.12 & 6.17 & 32.60 & 39.40 & 39.20 \\
& ESM-IF & 8.18 & 6.33 & 6.44 & 31.30 & 38.50 & 38.30\\
& \hspace{1em} + 1.2M AF2 data & 6.05 & 4.00 & 4.01 & 38.10 & 51.50 & 51.60 \\
& ProteinMPNN & 6.35 & 6.25 & 4.89 & 40.73 & 41.18 & 47.69 \\
& PiFold & 5.50 & 5.76 & 4.44 & 43.84 & 44.32 & 50.62 \\
& LM-Design & 5.66 & 5.52 & 4.01 & 42.84 & 43.69 & 55.68 \\
& Knowledge-Design & 5.47 & 5.23 & \underline{3.49} & 43.89 & 45.95 & \underline{60.38} \\
& Bridge-IF & \underline{5.17} & \underline{4.63} & 3.68 & \underline{50.00} & \underline{53.49} & 58.93 \\
\cmidrule(lr){2-8}
& \textbf{SymFold} & \textbf{4.26} & \textbf{3.91} & \textbf{3.22} & \textbf{55.81} & \textbf{58.20} & \textbf{62.23} \\
\bottomrule
\end{tabular}
\end{table*}

As shown in Table~\ref{tab:ts_results}, SymFold establishes new state-of-the-art results on both TS50 and TS500, which are more diverse than CATH. On TS50, it achieves a perplexity of 2.84 and recovery of 66.02\%, surpassing Knowledge-Design by 3.23 percentage points in recovery and reducing perplexity by 0.26. On the larger TS500, SymFold reaches 70.48\% recovery with the lowest perplexity of 2.74. These consistent improvements, with perplexity reductions of 8.4\% (TS50) and 4.2\% (TS500), demonstrate both scalability and robustness of the symmetric architecture in capturing sequence--structure relationships.

\begin{table}[ht]
\centering
\small
\setlength{\tabcolsep}{2pt} 
\caption{Results comparison on the TS50\&TS500 dataset. Some benchmarked results are quoted from~\citep{kwdesign}. The best and suboptimal results are labeled with bold and underline.}
\label{tab:ts_results}
\resizebox{\columnwidth}{!}{
\begin{tabular}{lcccc}
\toprule
\multirow{2}{*}{Model} & \multicolumn{2}{c}{TS50} & \multicolumn{2}{c}{TS500} \\
\cmidrule(lr){2-3} \cmidrule(lr){4-5}
& Perp. $\downarrow$ & Rec. \% $\uparrow$ & Perp. $\downarrow$ & Rec. \% $\uparrow$ \\ 
\midrule
StructGNN & 5.40 & 43.89 & 4.98 & 45.69 \\
GraphTrans & 5.60 & 42.20 & 5.16 & 44.66 \\
GVP & 4.71 & 44.14 & 4.20 & 49.14 \\
GCA & 5.09 & 47.02 & 4.72 & 47.74 \\
AlphaDesign & 5.25 & 48.36 & 4.93 & 49.23 \\
ProteinMPNN & 3.93 & 54.43 & 3.53 & 58.08 \\
PiFold & 3.86 & 58.72 & 3.44 & 60.42 \\
LM-Design & 3.50 & 57.89 & 3.19 & 63.65 \\
Knowledge-Design & \underline{3.10} & \underline{62.79} & \underline{2.86} & \underline{69.19} \\
\midrule
\textbf{SymFold} & \textbf{2.84} & \textbf{66.02} & \textbf{2.74} & \textbf{70.48} \\
\bottomrule
\end{tabular}
}
\end{table}

To further validate our model's generalization capability, we evaluated SymFold on the CASP15 and CASP16 datasets~\cite{casp16}, which provide challenging benchmarks with the latest experimentally determined protein structures. As shown in Table~\ref{tab:casp_results}, SymFold achieves 56.57\% and 52.18\% recovery rates on CASP15 and CASP16 respectively, outperforming ProteinMPNN (43.06\% and 39.32\%), PiFold (48.45\% and 47.10\%), and LM-Design (50.28\% and 48.64\%). We attempted to include Knowledge-Design and Bridge-IF, but encountered reproducibility issues with their released code. The consistent improvements over reproducible baselines demonstrate SymFold's strong performance on novel protein structures.
\begin{table}[ht]
\centering
\caption{Comparison of results on CASP15 and CASP16 datasets. The best and suboptimal results are labeled with bold and underline.}
\label{tab:casp_results}
\setlength{\tabcolsep}{4pt} 
\resizebox{\columnwidth}{!}{%
\begin{tabular}{lcccc}
\toprule
\multirow{2}{*}{Model} & \multicolumn{2}{c}{CASP15} & \multicolumn{2}{c}{CASP16} \\
\cmidrule(lr){2-3} \cmidrule(lr){4-5}
& Perp. $\downarrow$ & Rec. \% $\uparrow$ & Perp. $\downarrow$ & Rec. \% $\uparrow$ \\
\midrule
ProteinMPNN & 5.69 & 43.06 & 7.19 & 39.32 \\
PiFold & \underline{4.87} & 48.45 & \underline{5.66} & 47.10 \\
LM-Design & 5.12 & \underline{50.28} & 5.79 & \underline{48.64} \\
\midrule
\textbf{SymFold} & \textbf{3.90} & \textbf{56.57} & \textbf{4.50} & \textbf{52.18} \\
\bottomrule
\end{tabular}
}
\end{table}

\textbf{Q2. Are the various module designs in SymFold reasonable and valuable?}

\textbf{Ablations for our symmetric design}. We conduct ablation studies to validate our symmetric dual-path design by evaluating four configurations: (1) the complete SymFold with both streams, (2) SymFold without the PLM branch (w/o PLM), and (3) SymFold without the MPLM branch (w/o MPLM).

The results in Table~\ref{tab:dual_stream} demonstrate the synergistic benefits of our symmetric dual-path architecture. The full SymFold model achieves the best performance across all metrics, with perplexity of 3.23 and recovery rate of 63.11\% on the full test set. When we ablate the PLM branch (w/o PLM), keeping only the MPLM stream, performance drops significantly. Similarly, ablating the MPLM branch (w/o MPLM) while keeping only the PLM stream also degrades performance. These results confirm that both structural and sequence information are essential and complementary.
\begin{table}[ht]
\centering
\caption{Ablation results for the symmetric dual-path architecture. The full SymFold outperforms both single-path variants, confirming the complementary roles of PLM and MPLM.}
\label{tab:dual_stream}
\setlength{\tabcolsep}{2pt} 
\resizebox{\columnwidth}{!}{%
\begin{tabular}{lcccccc}
\toprule
\multirow{2}{*}{Variant} & \multicolumn{3}{c}{Perp. $\downarrow$} & \multicolumn{3}{c}{Rec. \% $\uparrow$} \\
\cmidrule(lr){2-4} \cmidrule(lr){5-7}
& Short & Single & All & Short & Single & All \\
\midrule
SymFold & \textbf{4.69} & \textbf{4.01} & \textbf{3.23} & \textbf{50.00} & \textbf{55.45} & \textbf{63.11} \\
\textit{w/o} PLM & 6.11 & 5.66 & 4.05 & 42.55 & 45.21 & 56.62 \\
\textit{w/o} MPLM & 5.51 & 4.73 & 3.61 & 44.02 & 50.89 & 59.56 \\
\bottomrule
\end{tabular}
}
\end{table}

Notably,on the CATH4.2 and CATH4.3 datasets, direct zero-shot evaluation of a pretrained MPLM yields perplexities of 6.50/6.27 and recovery rates of 42.03\%/40.98\%, demonstrating its inability to generalize effectively to inverse folding. Despite further fine-tuning, performance remains limited, with perplexities only improving to 4.82/4.68 and recovery rates to 48.46\%/48.73\%. This raw performance underscores that while MPLMs encode rich structural priors, effective adaptation—rather than direct application—is essential for unlocking their potential. By contrast, our symmetric dual-path design with joint training leverages the complementary strengths of PLMs and MPLMs, achieving significantly stronger results. For completeness, additional comparisons integrating the MPLM within another mainstream architecture are provided in the Appendix.

\textbf{Ablations for our proposed modules}. Table~\ref{tab:ablation} summarizes the ablation study of SymFold’s three core components: the Prior training strategy, Adaptive Synergistic Fusion (ASF), and Self-Correction (SC). The Prior strategy utilizes a frozen pre-trained encoder to extract stable structural features, which consistently lowers perplexity and aids sequence recovery. ASF optimizes decoding by dynamically integrating complementary signals from multiple expert modules, while SC mitigates error propagation through iterative refinement of intermediate predictions. The results validate the specific contribution of each module: the Prior establishes a robust representation foundation, ASF enhances signal reliability, and SC ensures output quality. When combined, these components operate synergistically to achieve the lowest perplexity and highest recovery across all categories, proving their collective essentiality for accurate and robust protein sequence generation.

\begin{table}[ht]
\centering
\caption{Performance comparison of SymFold when using ProteinMPNN versus PiFold as the structure encoder on CATH4.2.}
\label{tab:encoder_comparison}
\setlength{\tabcolsep}{2.5pt}
\resizebox{\columnwidth}{!}{%
\begin{tabular}{lcccccc}
\toprule
\multirow{2}{*}{Encoder} & \multicolumn{3}{c}{Perp. $\downarrow$} & \multicolumn{3}{c}{Rec. \% $\uparrow$} \\
\cmidrule(lr){2-4} \cmidrule(lr){5-7}
& Short & Single & All & Short & Single & All \\
\midrule
ProteinMPNN & 6.21 & 6.68 & 4.57 & 36.35 & 34.43 & 49.87 \\
\textit{with} SymFold & 5.02 & 4.56 & 3.34 & 48.96 & 52.28 & 61.45 \\
\midrule
PiFold & 6.04 & 6.31 & 4.18 & 39.84 & 38.53 & 51.66 \\
\textit{with} SymFold & 4.69 & 4.01 & 3.23 & 50.00 & 55.45 & 63.11\\
\bottomrule
\end{tabular}
}
\end{table}

\begin{table}[ht]
\centering
\caption{Ablation study of the three proposed components in SymFold: 
Prior (frozen pre-trained structure encoder), 
ASF (Adaptive Synergistic Fusion of predictions),
and SC (Self-Correction via iterative ). 
$\checkmark$ indicates the component is enabled, and $\times$ indicates it is disabled.}
\label{tab:ablation}
\setlength{\tabcolsep}{2pt}
\resizebox{\columnwidth}{!}{%
\begin{tabular}{ccccccccc}
\toprule
\multirow{2}{*}{Prior} & \multirow{2}{*}{ASF} & \multirow{2}{*}{SC} & \multicolumn{3}{c}{Perp. $\downarrow$} & \multicolumn{3}{c}{Rec. \% $\uparrow$} \\
\cmidrule(lr){4-6} \cmidrule(lr){7-9}
& & & Short & Single & All & Short & Single & All \\
\midrule
$\times$ & $\times$ & $\times$ & 4.98 & 4.40 & 3.46 & 49.52 & 53.00 & 61.26 \\
$\times$ & $\checkmark$ & $\checkmark$ & 4.92 & 4.32 & 3.40 & 48.56 & 53.26 & 61.85 \\
$\checkmark$ & $\times$ & $\checkmark$ & 4.83 & 4.20 & 3.35 & 49.75 & 54.52 & 62.31 \\
$\checkmark$ & $\checkmark$ & $\times$ & 4.80 & 4.18 & 3.32 & 49.81 & 54.60 & 62.43 \\
$\checkmark$ & $\checkmark$ & $\checkmark$ & \textbf{4.69} & \textbf{4.01} & \textbf{3.23} & \textbf{50.00} & \textbf{55.45} & \textbf{63.11} \\
\bottomrule
\end{tabular}
}
\end{table}

\begin{figure*}[h]
    \centering
    \includegraphics[width=1.0\textwidth]{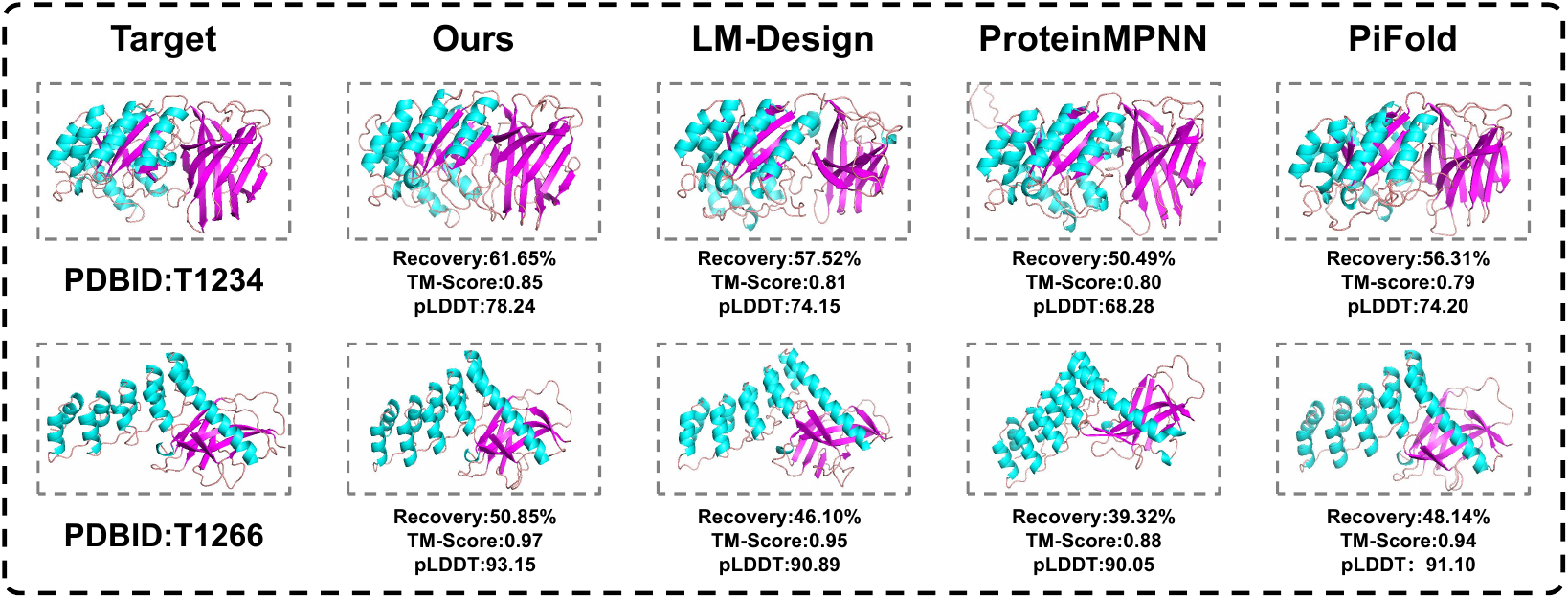}
    \caption{Comparison of sequence recovery, structure prediction confidence (pLDDT), and structural similarity (TM-score)~\cite{tmalign} for sequences generated by SymFold, LM-Design, ProteinMPNN, and PiFold. The TM-score ranges from 0 to 1 (higher values indicate better structural similarity), while pLDDT ranges from 0 to 100 (higher values indicate greater prediction confidence).}
    \label{fig:figure3}
\end{figure*}

\textbf{Q3. Is SymFold model-agnostic?}

We evaluated the SymFold framework's compatibility with different foundation models by integrating two distinct structure encoders: ProteinMPNN and PiFold. These experiments were conducted under controlled conditions, utilizing the same MPLM and PLM components while maintaining consistent training strategies. As shown in Table~\ref{tab:encoder_comparison}, our framework consistently enhances the performance of both encoders across all evaluated categories. SymFold achieves substantial improvements in sequence recovery for both ProteinMPNN (from 49.87\% to 61.45\%) and PiFold (from 51.66\% to 63.11\%), with corresponding perplexity reductions. Notably, when equipped with the more advanced PiFold encoder, SymFold achieves superior performance across all metrics, indicating that better structure encoders provide more precise prior conditions for the synergistic generation process. These results confirm that the SymFold architecture is fundamentally model-agnostic while scaling its performance with the quality of underlying components.

\textbf{Q4. Are the generated sequences biologically plausible?}

To evaluate the biological plausibility of sequences generated by SymFold, we conducted a validation experiment using AlphaFold3~\cite{alphafold3} for structure prediction. We selected two proteins from CASP16 with varying lengths: T1234 (377 residues) and T1266 (295 residues). For each target structure, we generated sequences using SymFold, LM-Design, PiFold, and ProteinMPNN, then used AlphaFold3 to predict structures from these designed sequences.

As shown in Figure~\ref{fig:figure3}, SymFold-generated sequences exhibit stronger consistency with the intended target structures across recovery, pLDDT, and TM-score. For T1234, our method achieves a sequence recovery of 61.65\%, higher than LM-Design (57.52\%), ProteinMPNN (50.49\%), and PiFold (56.31\%). For T1266, SymFold attains 50.85\% compared to 46.10\%, 39.32\%, and 48.14\% for the respective baselines. AlphaFold3 predictions suggest that sequences designed by SymFold yield more confident folding models (pLDDT: 78.24 for T1234, 93.15 for T1266) with better structural alignment to the native backbone (TM-scores of 0.85 and 0.97, respectively).

Additional visual comparisons on four more CASP16 targets are provided in Appendix. These results provide preliminary in silico evidence that our dual-stream architecture can improve not only recovery metrics but also the predicted structural plausibility of designed sequences. While AlphaFold-derived scores cannot substitute for experimental verification, they offer encouraging indications that SymFold generated sequences are more compatible with the desired folds.

\section{Conclusion}
We introduce SymFold, a novel symmetric dual-path framework for protein inverse folding that addresses the limitations of serial architectures by fostering continuous interaction between structural constraints and Sequence Evolution knowledge. Our Adaptive Synergistic Fusion and self-correction mechanisms enhance sequence design, achieving state-of-the-art performance across benchmarks. Despite these advances, challenges remain. Our evaluations primarily rely on in silico metrics, and the limited extent of wet-lab validation leaves the functional viability of the generated sequences unverified. Future work will explore stronger foundation models to push the boundaries of computational protein design.

\section{Supplementary Material}

\subsection{Comparative Architecture Designs with MPLM}
\label{Alm_design}
To further investigate the role of cross-modal integration of MPLMs in protein inverse folding, we replaced the original protein language model (PLM) in the LM-Design framework with an MPLM. By systematically evaluating two representative adapter schemes within this framework, this study thoroughly examines how different connection strategies between multimodal protein language models and structure encoders affect downstream recovery performance, thereby providing clearer insights into the mechanisms of synergy across modalities.We evaluated two representat:

(1) Structure-Aware Adapter (original LM-Design implementation): 
This variant integrates high-dimensional embeddings from a protein structure encoder through lightweight adapter modules. 
Cross-modal interaction is realized via multi-head attention, explicitly mixing coordinate-based information with sequence embeddings.  

(2) Parameter-Tuning Adapter (modified version):
Here, the attention-based fusion is removed. The adapter reduces to simple trainable MLP layers, 
leaving structural sensitivity to be modeled implicitly by the MPLM itself without direct feature injection.  

Both architectures were trained under identical setups (datasets, losses, optimizers, and hyperparameters). 
We further controlled two experimental factors:  
(i) \textit{Training Strategy} – training the structure encoder from scratch versus freezing pretrained encoder parameters;  
(ii) \textit{Input Modality} – full multimodal input (amino-acid sequence + 3D coordinates) versus sequence-only input.  

\begin{table}[h] 
\centering
\setlength{\tabcolsep}{4pt} 
\resizebox{\columnwidth}{!}{
\begin{tabular}{llcc}
\toprule
\textbf{Training Strategy} & \textbf{Architecture} & \textbf{Full Input} & \textbf{Seq. Only} \\ 
\midrule
\multirow{2}{*}{From Scratch} 
    & Struct-Aware Adapter  & 52.46 & 53.68 \\
    & Param-Tuning Adapter  & \textbf{53.60} & 53.12 \\
\midrule
\multirow{2}{*}{Pretrained + Frozen} 
    & Struct-Aware Adapter  & 53.95 & \textbf{54.56} \\
    & Param-Tuning Adapter  & \textbf{55.33} & 54.01 \\
\bottomrule
\end{tabular}
}
\caption{Recovery performance comparison of adapter architectures with MPLM backbone under different training strategies and input modalities.}
\label{tab:LM-Design-with-mplm}
\end{table}

This counterintuitive result reveals a phenomenon we term the \textbf{modality paradox}: providing richer multimodal inputs does not automatically translate into better performance and may even lead to degradation due to issues such as feature redundancy and optimization conflicts. In contrast, the Parameter-Tuning Adapter can more stably leverage the inherently structure-aware capacity of MPLMs, indicating that effective cross-modal adaptation requires carefully designed fusion mechanisms. Based on these experiments, we preliminarily conclude that when integrating MPLMs (e.g., ESM3) into protein inverse folding frameworks, it is preferable to preserve their original architecture without substantial modifications. Otherwise, as observed in the LM‑Design framework introduced with MPLMs, performance may even decline. This finding provides strong foundational support for the SymFold framework proposed in this paper.

\subsection{Iterative Inference Analysis}
\label{Aiter}
\label{sec:iterative_inference}
We analyzed the impact of iterative refinement steps ($T$) on DualFold's performance with and without Self-Correction. Table~\ref{tab:iterative_inference} shows the results on CATH4.2.

\begin{table}[h]
\centering
\begin{tabular}{ccc}
\toprule
\multirow{2.3}{*}{$T$} & \multicolumn{2}{c}{Recovery(\%) $\uparrow$} \\
\cmidrule(lr){2-3}
& With Self-Correction & Without Self-Correction \\
\midrule
1 & 62.25 & 60.57 \\
2 & \textbf{63.11} & 61.19 \\
3 & 62.96 & 62.23 \\
4 & 62.96 & \textbf{62.43} \\
5 & 63.10 & 62.42 \\
\bottomrule
\end{tabular}
\caption{Impact of iterative refinement steps ($T$) on sequence recovery rate (\%).}
\label{tab:iterative_inference}
\end{table}
The results clearly demonstrate the effectiveness of the Self-Correction mechanism. With Self-Correction, the model achieves superior performance across all iteration counts, with the most significant improvements at lower iterations. Notably, Self-Correction enables the model to reach peak performance (63.11\%) early in the refinement process. In contrast, without Self-Correction, the model shows slower convergence and lower overall performance, never reaching the accuracy achieved by Self-Correction even with more iterations.

\subsection{ CASP15 and CASP16 Protein Details}
\label{Acasp}
For the purpose of ensuring reproducibility, this appendix summarizes the specific protein targets used in our experiments from the CASP15 and CASP16 datasets.  
Tables~\ref{tab:casp15-list} and~\ref{tab:casp16-list} provide the official release names and dates of each protein.  
This collection allows future studies to easily identify and cross-reference proteins with their corresponding release times.

\vspace{0.5em}
\begin{table*}[h]
\centering
\setlength{\tabcolsep}{8pt} 
\renewcommand{\arraystretch}{1.1}
\small 
\begin{tabular}{|l|c|l|c|l|c|}
\hline
\textbf{Protein Name} & \textbf{Release Time} & 
\textbf{Protein Name} & \textbf{Release Time} & 
\textbf{Protein Name} & \textbf{Release Time} \\ \hline
T1104-D1    & 2022-07-11 & T1106s1-D1  & 2022-06-06 & T1106s2-D1  & 2022-06-06 \\ \hline
T1109-D1    & 2022-05-31 & T1119-D1    & 2022-06-08 & T1120-D1    & 2022-07-14 \\ \hline
T1120-D2    & 2022-07-14 & T1121-D1    & 2022-06-08 & T1121-D2    & 2022-06-08 \\ \hline
T1123-D1    & 2022-05-19 & T1124-D1    & 2022-07-16 & T1129s2-D1  & 2022-07-05 \\ \hline
T1133-D1    & 2022-07-18 & T1137s1-D1  & 2022-06-22 & T1137s1-D2  & 2022-06-22 \\ \hline
T1137s2-D1  & 2022-05-30 & T1137s2-D2  & 2022-05-30 & T1137s3-D1  & 2022-05-31 \\ \hline
T1137s3-D2  & 2022-06-22 & T1137s4-D1  & 2022-06-22 & T1137s4-D2  & 2022-06-22 \\ \hline
T1137s4-D3  & 2022-06-22 & T1137s5-D1  & 2022-06-22 & T1137s5-D2  & 2022-06-22 \\ \hline
T1137s6-D1  & 2022-06-22 & T1137s6-D2  & 2022-06-22 & T1137s7-D1  & 2022-06-01 \\ \hline
T1137s8-D1  & 2022-06-01 & T1137s9-D1  & 2022-06-01 & T1139-D1    & 2022-06-01 \\ \hline
T1145-D1    & 2022-08-18 & T1145-D2    & 2022-08-18 & T1150-D1    & 2022-06-11 \\ \hline
T1157s1-D1  & 2022-09-01 & T1157s1-D2  & 2022-09-01 & T1157s1-D3  & 2022-09-01 \\ \hline
T1157s2-D1  & 2022-09-01 & T1157s2-D2  & 2022-09-01 & T1157s2-D3  & 2022-09-01 \\ \hline
T1170-D1    & 2022-07-28 & T1170-D2    & 2022-07-28 & T1180-D1    & 2022-08-24 \\ \hline
T1187-D1    & 2022-08-05 & T1188-D1    & 2022-08-05 & T1194-D1    & 2022-08-05 \\ \hline
\end{tabular}
\caption{CASP15 protein names and release times.}
\label{tab:casp15-list}
\end{table*}

\begin{table*}[h]
\centering
\setlength{\tabcolsep}{8pt}
\renewcommand{\arraystretch}{1.1}
\small
\begin{tabular}{|l|c|l|c|l|c|}
\hline
\textbf{Protein Name} & \textbf{Release Time} & 
\textbf{Protein Name} & \textbf{Release Time} & 
\textbf{Protein Name} & \textbf{Release Time} \\ \hline
T1237    & 2024-07-20 & T1206    & 2024-07-17 & T1234-D1   & 2024-09-09 \\ \hline
T1276    & 2024-08-19 & T1276-D1 & 2024-08-19 & T1272s3    & 2024-07-29 \\ \hline
T1279-D1 & 2024-07-25 & T1214v1  & 2024-08-12 & T1272s9    & 2024-11-14 \\ \hline
T1212    & 2024-07-04 & T1272s1  & 2024-07-29 & T1259      & 2024-07-25 \\ \hline
T1235    & 2024-07-04 & T1279    & 2024-07-20 & T1298-D1   & 2024-08-16 \\ \hline
T1274-D1 & 2024-09-03 & T1201-D1 & 2024-09-24 & T1298-D2   & 2024-08-16 \\ \hline
T1272s6-D1 & 2024-11-14 & T1240-D1 & 2024-08-02 & T1201    & 2024-05-19 \\ \hline
T1212-D1 & 2024-07-04 & T1237-D1 & 2024-07-20 & T1266-D1 & 2024-07-02 \\ \hline
T1235-D1 & 2024-09-09 & T1234    & 2024-07-04 & T1299-D1 & 2024-09-11 \\ \hline
T1266    & 2024-07-02 & T1272s7  & 2024-07-29 & T1272s8-D1 & 2024-11-14 \\ \hline
T1214v2  & 2024-08-12 & T1272s2-D1 & 2024-08-03 & T1299    & 2024-09-11 \\ \hline
T1272s8  & 2024-11-14 & T1274    & 2024-08-14 & T1298     & 2024-08-16 \\ \hline
T1272s6  & 2024-11-14 & T1240    & 2024-08-01 & T1210     & 2024-07-15 \\ \hline
T1272s2  & 2024-11-14 & T1214-D1 & 2024-09-12 & T1272s4   & 2024-07-29 \\ \hline
T1206-D1 & 2024-07-17 & T1259-D1 & 2024-07-25 & T1210-D1  & 2024-09-24 \\ \hline
T1272s9-D1 & 2024-11-14 & T1240-D2 & 2024-08-02 & T1279-D2 & 2024-07-25 \\ \hline
T1272s5 & 2024-07-29 & T1214    & 2024-07-10 &           &            \\ \hline
\end{tabular}
\caption{CASP16 protein names and release times.}
\label{tab:casp16-list}
\end{table*}

\subsection{Ablation on ASF Module Components}

\begin{figure}[h]
    \centering
    \includegraphics[width=0.85\linewidth]{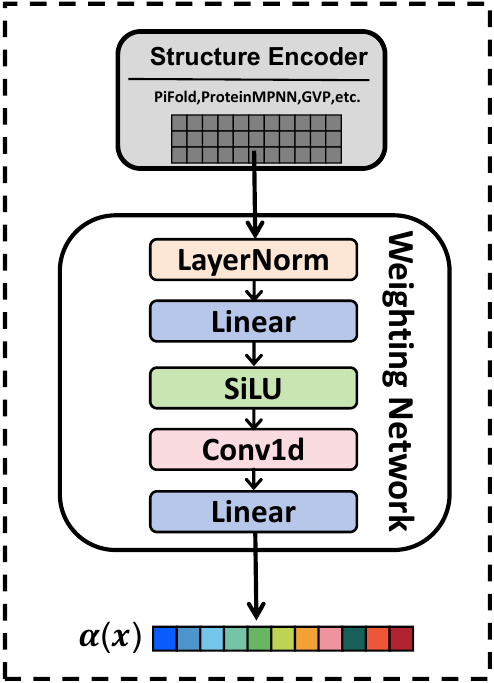}
    \caption{This module takes features extracted by structure encoders as input, processes them through a sequential layer comprising Layer Normalization, Linear transformation, Sigmoid Linear Unit , and 1D Convolution , and finally outputs position-wise adaptive normalized weighting coefficients $\alpha$(x).}
    \label{fig:figure5}
\end{figure}

To effectively integrate the predictive capabilities of MPLMs and PLMs, we propose a residue-wise weighting strategy. We constructed a lightweight ResidueWeightingNetwork that utilizes features extracted by the encoder to capture local sequence dependencies through hybrid layers comprising Sigmoid Linear Units (SiLU) and local 1D convolutions (Conv1d). This network outputs normalized weights for each model at every position within the sequence. This residue-wise weighting mechanism allows the model to dynamically adjust the contribution of each component based on the specific contextual environment of the amino acids, thereby achieving enhanced feature fusion at the local level.The specific architecture diagram is shown in Figure \ref{fig:figure5}.

To further validate the effectiveness of the Adaptive Synergistic Fusion (ASF) module in our SymFold framework, we conducted ablation studies on its key components. The ASF module, as described in the main paper (Equation 4), dynamically integrates sequence priors from the PLM ($f_s$) and structural priors from the MPLM ($f_m$) using a residue-wise gating vector $\alpha(x) \in \mathbb{R}^L$, where $L$ is the sequence length. This allows the model to balance evolutionary and structural signals at a per-residue (token) level, conditioned on the structural embeddings from the frozen structure encoder $f_e^*$.

We compare the proposed token-level adaptive fusion against several variants:
\begin{itemize}
    \item \textbf{Average Fusion}: A simple uniform averaging of the logits from $f_s$ and $f_m$, where each model contributes equally without any adaptive weighting ($\alpha(x) = 0.5$ for all residues).
    \item \textbf{Model-Leval Fusion}: A variant employing two learnable global scalars ($\alpha, \beta$) that assign fixed weights to the outputs of the PLM and MPLM, applied uniformly across all residues.
    \item \textbf{Sequence-Level Mixing}: Adaptive weighting at the sequence level, where all tokens in a sequence share a single weight vector derived from global structural features.
    \item \textbf{Token-Level Mixing (Ours)}: The proposed residue-wise adaptive fusion, where each token has an independent weight computed via a lightweight projection network on per-residue embeddings.
\end{itemize}

These variants were evaluated on the CATH4.2 benchmark, measuring sequence recovery rate (\%) as the primary metric. The results are summarized in Table~\ref{tab:asf_ablation}.

\begin{table}[h]
\centering
\begin{tabular}{lc}
\toprule
\textbf{Fusion Variant} & \textbf{Recovery (\%)} \\
\midrule
Average Fusion & 62.31 \\
Model-leval & 62.15 \\
Sequence-Level Mixing & 62.35 \\
Token-Level Mixing (Our ASF) & \textbf{63.11} \\
\bottomrule
\end{tabular}
\caption{Ablation on ASF module components. Performance is reported as sequence recovery rate (\%) on CATH4.2.}
\label{tab:asf_ablation}
\end{table}

The results demonstrate that finer-grained adaptive fusion yields superior performance. The token-level mixing in our ASF outperforms coarser variants by up to 0.96\%, highlighting the importance of residue-specific balancing. This is particularly beneficial for proteins where structural constraints vary across residues (e.g., core vs. surface regions). Coarser fusions, such as sequence-level or global weighting, fail to capture these local variations, leading to suboptimal integration of priors. These findings corroborate the rationale for ASF's design, as coarser alternatives compound errors in structurally heterogeneous regions, while token-level adaptation enhances compatibility with both evolutionary and geometric constraints.

\subsection{Additional Folding Results with Alphafold3}
\label{Avisual}

\begin{figure*}[t]
    \centering
    \includegraphics[width=\linewidth]{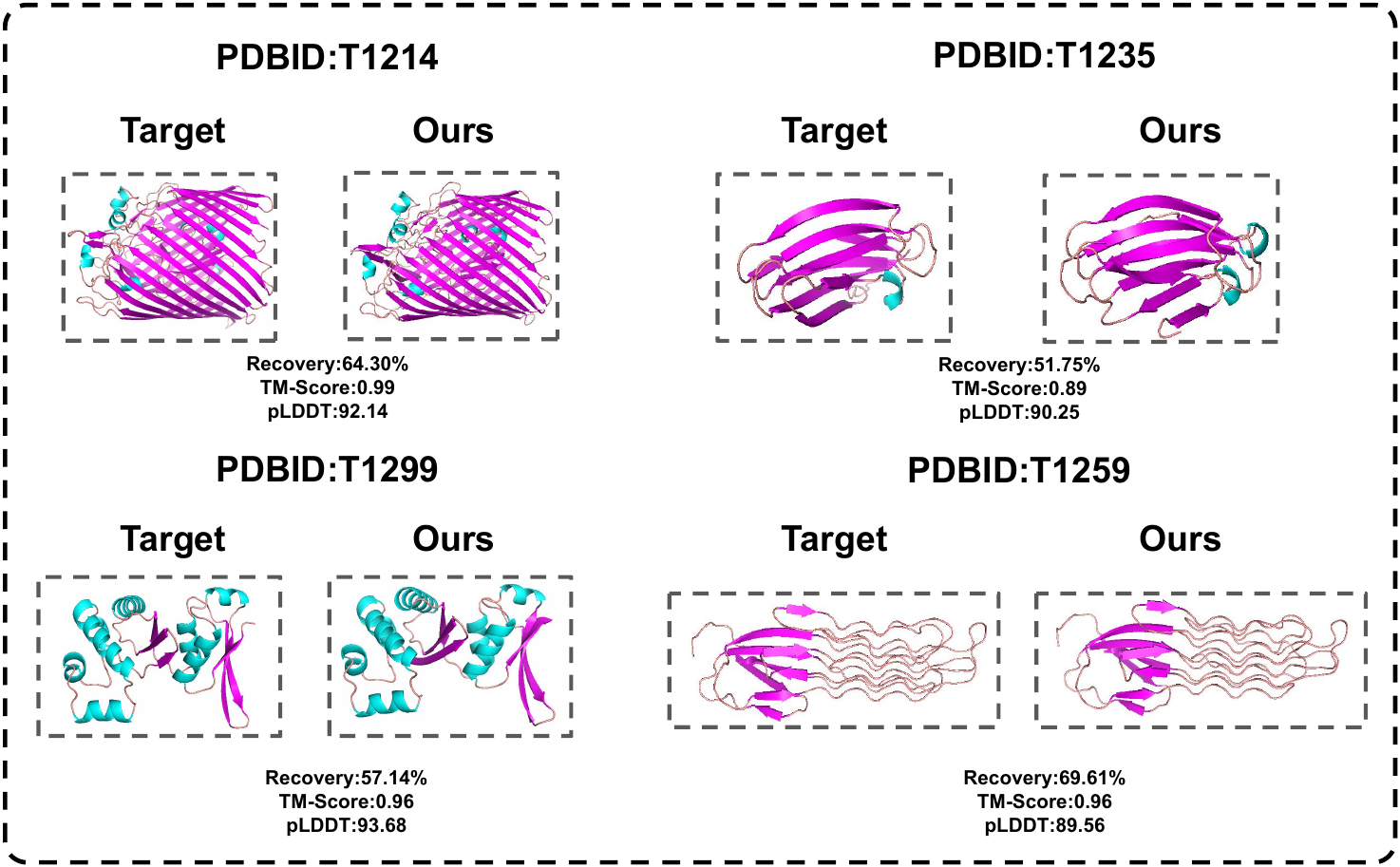}
    \caption{AlphaFold3-predicted structures for sequences designed by DualFold 
    on additional CASP16 targets (T1214, T1235, T1299, and T1259). Each row 
    shows an overlay of the target structure and the predicted structure 
    from the DualFold-designed sequence. Across all cases, the strong 
    alignment provides further support for the biological plausibility of 
    DualFold outputs.}
    \label{fig:figure4}
\end{figure*}
To further validate the biological plausibility of DualFold-generated sequences, 
we extended our AlphaFold3 analysis to four additional CASP16 
targets: T1214(length=652), T1235(length=114), T1299(length=68), and T1259(length=204). For each case, we compared the 
AlphaFold3-predicted structure of the sequence produced by DualFold against the 
native target backbone.

As illustrated in Figure~\ref{fig:figure4}, the predicted structures of 
all four designed sequences align closely with their respective native folds.
Visually, the confidence levels (pLDDT) remain consistently high, and the 
structural overlays show strong backbone agreement. These results further 
reinforce that DualFold is capable of generating sequences with not only improved 
design metrics but also high in silico structural plausibility across targets of 
different lengths and topologies.

\subsection{LoRA Configuration for MPLM and PLM Fine-tuning}
\label{Aora}
Both models employ identical LoRA hyperparameters to ensure balanced adaptation. We use a rank of 8, which provides a good balance between parameter efficiency and model expressivity. The scaling factor is set to 32, offering sufficient learning capacity for the inverse folding task while maintaining stability. A dropout rate of 0.1 is applied to the LoRA layers for regularization, and we disable bias adaptation to further reduce the parameter count.

This configuration results in approximately 0.1\% trainable parameters relative to the full model size, significantly reducing computational overhead while maintaining the models' ability to adapt to the inverse folding task. The low rank ensures efficient adaptation, while the relatively high scaling factor compensates for the reduced parameter space, allowing the models to learn task-specific patterns effectively.

The LoRA adaptation is implemented using the Parameter Efficient Fine-Tuning (PEFT) framework. During initialization, we load the pretrained ESM-3 and ESM-C models, configure the LoRA adapters with the specified hyperparameters, and freeze all base model parameters. Only the LoRA weights are updated during training, which prevents catastrophic forgetting of the pretrained knowledge while allowing task-specific adaptation.

The symmetric configuration across both models ensures balanced adaptation of sequence and structural knowledge during training. By targeting both attention and feed-forward components, we enable the models to adapt their representation learning capabilities while preserving the fundamental understanding encoded in the pretrained weights.

\subsection{Mathematical Formulations for Evaluation Metrics}
\label{Ametrics}
\textbf{Perplexity}

Perplexity is a fundamental metric for evaluating the quality of language model predictions, defined as the exponential of cross-entropy loss. In protein sequence design tasks, we utilize perplexity to assess the model's accuracy in predicting authentic amino acid sequences.

Given a protein sequence $\mathbf{s} = \{s_1, s_2, \ldots, s_\ell\}$, where $s_i$ represents the amino acid at position $i$ and $\ell$ is the sequence length. The model outputs logits denoted as $\mathbf{z} = \{\mathbf{z}_1, \mathbf{z}_2, \ldots, \mathbf{z}_\ell\}$, where $\mathbf{z}_i \in \mathbb{R}^{|\mathcal{V}|}$ and $|\mathcal{V}|$ is the vocabulary size (20 standard amino acids).

First, we compute the log probability distribution at each position:
\begin{equation}
p_i(k) = \frac{\exp(z_{i,k})}{\sum_{j=1}^{|\mathcal{V}|} \exp(z_{i,j})}
\end{equation}

where $z_{i,k}$ represents the logit value for predicting amino acid $k$ at position $i$.

The log-likelihood for the target amino acid is:
\begin{equation}
\log p_i(s_i) = \log \left( \frac{\exp(z_{i,s_i})}{\sum_{j=1}^{|\mathcal{V}|} \exp(z_{i,j})} \right) = z_{i,s_i} - \log \sum_{j=1}^{|\mathcal{V}|} \exp(z_{i,j})
\end{equation}

\textbf{Global Perplexity Calculation}

Considering the presence of special tokens such as padding, class (cls), and end-of-sequence (eos) tokens in protein sequences, we define a valid position mask:
\begin{equation}
\mathcal{M}_i = \mathbb{1}[s_i \neq \text{pad}] \cdot \mathbb{1}[s_i \neq \text{cls}] \cdot \mathbb{1}[s_i \neq \text{eos}] \cdot c_i
\end{equation}

where $c_i$ is the coordinate mask used to identify structurally valid positions, and $\mathbb{1}[\cdot]$ is the indicator function.

The global average negative log-likelihood is:
\begin{equation}
\text{NLL}_{\text{global}} = -\frac{\sum_{i=1}^{\ell} \mathcal{M}_i \cdot \log p_i(s_i)}{\sum_{i=1}^{\ell} \mathcal{M}_i}
\end{equation}

The final global perplexity is defined as:
\begin{equation}
\text{PPL}_{\text{global}} = \exp(\text{NLL}_{\text{global}})
\end{equation}

\textbf{Sequence Recovery Rate}

The sequence recovery rate measures the degree of matching between generated sequences and target sequences at the amino acid level, serving as a direct indicator for evaluating the accuracy of protein sequence design.

Given a predicted sequence $\hat{\mathbf{s}} = \{\hat{s}_1, \hat{s}_2, \ldots, \hat{s}_\ell\}$ and a target sequence $\mathbf{s} = \{s_1, s_2, \ldots,s_\ell\}$, the sequence recovery rate is defined as the proportion of correct predictions at valid positions.

The correctness indicator function for individual positions:
\begin{equation}
\delta_i = \mathbb{1}[\hat{s}_i = s_i] \cdot \mathcal{M}_i
\end{equation}

where $\mathcal{M}_i$ is the valid position mask defined previously.

\textbf{Sequence-level Recovery Rate}

For a single sequence, the recovery rate is calculated as:
\begin{equation}
\text{Recovery}_{\text{seq}} = \frac{\sum_{i=1}^{\ell} \delta_i}{\sum_{i=1}^{\ell} \mathcal{M}_i}
\end{equation}

\textbf{Dataset-level Statistics}

For a test set containing $N$ sequences, we compute the recovery rate for each sequence and then calculate the median as the final evaluation metric:
\begin{equation}
\text{Recovery}_{\text{median}} = \text{median}\{\text{Recovery}_{\text{seq}}^{(1)}, \ldots, \text{Recovery}_{\text{seq}}^{(N)}\}
\end{equation}

The use of median instead of mean reduces the impact of outliers and better reflects the model's overall performance.

\section{Acknowledgments}
This work was supported by the Strategic Priority Research
Program of the Chinese Academy of Sciences under Grant
No. XDA0460205

\bibliographystyle{named}
\bibliography{ref}

\end{document}